# Do readers prefer AI-generated Italian short stories?

**Michael Farrell**
IULM University
Milan
Italy
michael.farrell@iulm.it

## Abstract

This study investigates whether readers prefer AI-generated short stories in Italian over one written by a renowned Italian author. In a blind setup, 20 participants read and evaluated three stories, two created with ChatGPT-4o and one by Alberto Moravia, without being informed of their origin. To explore potential influencing factors, reading habits and demographic data, comprising age, gender, education and first language, were also collected. The results showed that the AI-written texts received slightly higher average ratings and were more frequently preferred, although differences were modest. No statistically significant associations were found between text preference and demographic or reading-habit variables. These findings challenge assumptions about reader preference for human-authored fiction and raise questions about the necessity of synthetic-text editing in literary contexts.

## 1 Introduction

In a recent study, Farrell (2026) asked 69 professional translators to distinguish between AI-generated short stories in Italian and a human-written text. Nearly as many participants mistook the synthetic texts (SynTs) for the human-authored text (HAT) as correctly identified them. This may suggest that the participants perceived the SynTs as better written, more *human-like* and maybe even preferred them.

Similarly, a study by Zhang and Gosline (2023) found that advertising content generated in English by AI, as well as human-created English advertising content augmented by AI (i.e., automatically edited), was perceived as higher quality than content produced solely by human experts. Likewise, a study by Porter and Machery (2024) revealed that AI-generated English poetry is indistinguishable from human-written verse and is rated more favourably.

To test whether some people prefer reading Italian SynT, it was decided to conduct an experiment on a group of volunteer readers at a public library using the same short stories used in Farrell's previous experiment

## 2 Aims

The principal objective of this experiment was:

- To determine whether people prefer reading short stories in Italian generated by AI or a short story written by an eminent human author.

There were also two secondary aims:

- To determine whether age, gender, educational background, first language, reading frequency, kind of things people normally read and things they prefer to read affect their preference or otherwise for AI-generated text.
- To shed light on the need for synthetic-text editing (STE), which aims to make the text "more engaging" and give it "a more human voice" (Farrell, 2025b). If people prefer reading unedited SynT, this form of editing might be unnecessary, at least for the kind of texts examined in this study.

## 3 Method

The experiment was conducted at the public library (*Biblioteca Civica "F. Pezza"*) in Mortara, Pavia, Italy and the participants were recruited through leaflets distributed on-site and a local newspaper article promoting the event a few days before it took place. The study was presented as an investigation into the factors influencing readers' preferences for particular texts.

The participants were asked to read three short stories in Italian of similar length (1807, 1626 and 1331 words). It was estimated that it would take about 24 minutes to read all three texts, considering an average reading speed of about 200 words per minute. Since they were also asked to answer questions and write assessments, in practice they were allowed about 45 minutes.

The participants were not told that any of the texts were generated by AI to prevent them from expressing their opinions on the basis that they might have suspected the texts to be synthetic. The titles were removed and replaced with geometric shapes (an oval, a hexagon and a five-pointed star) so that the texts could be identified during the experiment. These shapes were chosen since they lack a natural order and therefore do not suggest a ranking.

Each participant received an envelope containing the three short stories in random order, a questionnaire, a chart showing a few example opinions one might want to express when criticizing a short story and a pen. At the end of each short story, there was a form to be filled in with a score representing how much they had liked the story from 0 (I did not like it at all) to 10 (I liked it very much), together with a space in which they could explain the reason for the score they assigned (in free form). The participants were also encouraged to underline or mark the parts of the text which had informed their opinions. Moreover, they were asked whether they recognized the story or thought they had already read it, or whether they believed they could identify the style of a specific author.

The opinion expressed had to be personal. Therefore, consultation with other people, including other participants in the experiment, was not permitted. Internet consultation via laptop, smartphone or other devices was also prohibited.

The questionnaire asked for some demographic data (age, gender, educational background and first language), information on reading habits (reading frequency on a scale from 1 to 10 and kinds of texts read most frequently) and personal preferences (in the case of a free choice, kinds of text read most willingly). Lastly, consent for the use and publication of anonymized data was requested on a separate form to ensure participant anonymity.

At the end of the experiment, the participants placed all the materials received back into the envelope, sealed it and returned it to the researcher.

After the envelopes were returned, there was a short debriefing to satisfy the curiosity of the participants regarding the names of the authors and the titles of the stories, during which it was revealed that two of the texts were generated by AI.

## 3.1 Texts used

The two AI-written texts used in this experiment were generated by ChatGPT-4o for two earlier studies (Farrell, 2025a; 2026), which examined whether postgraduate translation students and professional translators, respectively, were able to distinguish SynT from HAT. The stories were created using prompts designed to emulate the plot and style of the human-written text used in the same experiments and were not machine translations of an existing source text. Full details of the prompts and prompt-engineering techniques are provided in the first cited paper.

The texts were labelled with the same geometric shapes used in Farrell's second experiment: the SynT marked with a hexagon corresponded to ST5 in his first study, while the one marked with a five-pointed star corresponded to ST2. The human-written text was once again Alberto Moravia's short story *L'incosciente* (The Reckless Man), from *Racconti romani* (Roman Tales, 1954), and was marked with an oval. It was chosen for its brevity and because it predates machine translation and AI, which ensures that these technologies could not have played any role in its creation.

As in his second experiment, involving professional translators, the three short stories were not divided into consecutive short extracts but were presented in full to the participants.

All three texts were analysed using the Plagramme AI detector[1] to determine whether any objectively measurable differences existed between them.

[1] www.plagramme.com

| Text | Kind | Length (words) | Mean score | Standard deviation | 1st place rankings | AI detector score |
|---|---|---|---|---|---|---|
| **Oval** | Human | 1807 | 6.83 | ± 1.71 | 6 | 17% |
| **Hexagon** | Synthetic | 1626 | 7.33 | ± 2.32 | 9 | 94% |
| **Star** | Synthetic | 1331 | 7.42 | ± 1.93 | 7 | 83% |

Table 1: Scores assigned to short stories

### 3.2 Statistical analyses

Two exact tests were used to analyse the demographic and reading-habits data. The Fisher–Freeman–Halton exact test was computed using Dr. Daniel Soper's calculator[2], while Fisher's exact test was computed using Jeremy Stangroom's calculator[3], except for the comparison marked with an asterisk in Table 2, for which Navendu Vasavada's calculator[4] was used since it allows for zero cell values, unlike Stangroom's. Spearman's rank correlation coefficient (ρ) was used to examine correlations with the results of the previous experiments, while a scatterplot was used to compare the findings with AI detector data.

## 4 Results

### 4.1 Scores

Twenty volunteers chose to attend the public library for the experiment reported in this study. Two were excluded for not rating all three short stories. Table 1 shows the mean score and number of first-place[5] rankings assigned to each short story.

### 4.2 Reasoning given for the scores assigned to the oval text (HAT)

#### 4.2.1 Positive comments (common)

The readers frequently praised the narrative flow and dialogue, describing it as engaging, fluid and psychologically credible. Several noted the humour, original narrative perspective (first-person) and well-developed characters. Some appreciated the depth of themes like conscience and awareness, which were handled with sensitivity.

#### 4.2.2 Negative comments (common)

Many criticized the slow or unengaging pacing and overly mechanical structure. A few found the language either too simplistic or mildly irritating, despite being appropriate for the characters. Some found the text prolix, confusing or monotonous, lacking suspense or brilliance.

#### 4.2.3 Neutral comments (occasional)

A couple of comments were descriptive rather than evaluative, e.g., noting the main character's redemption or expressing curiosity about an unfamiliar word.

### 4.3 Reasoning given for the scores assigned to the hexagon text (SynT)

#### 4.3.1 Positive comments (common)

Frequently described as well-written, clear and emotionally resonant. Multiple readers appreciated the structure, fluid style and the depth of emotional development. The text was often seen as engaging, with a strong sense of narrative progression.

#### 4.3.2 Negative comments (occasional to common)

Some readers felt it lacked originality, describing it as clichéd, banal or too moralistic. A few pointed out superficial character psychology or found its themes predictable and unexciting.

#### 4.3.3 Neutral comments (occasional)

A few remarks described the story's structure or moral development in neutral or factual terms, without clear praise or criticism.

### 4.4 Reasoning given for the scores assigned to the star text (SynT)

#### 4.4.1 Positive comments (common)

Many found the writing fluid, immersive and emotionally effective. Several highlighted the moral depth or psychological insight. The conclusion and introspective tone were often praised.

[2] www.danielsoper.com/statcalc/calculator.aspx?id=58
[3] www.socscistatistics.com/tests/
[4] astatsa.com/FisherTest/
[5] Joint first place was permitted; therefore, the total exceeds the number of participants.

| Demographic | Oval (p value / test) | Hexagon (p value / test) | Star (p value / test) |
|---|---|---|---|
| **Age group** | .413 / FFH | .242 / FFH | 1.000 / FFH |
| **Gender** | .569 / F | 1.000 / F | .057 / F* |
| **Educational level** | .510 / FFH | .615 / FFH | 1.000 / FFH |
| **First language** | .200 / F | .242 / F | 1.000 / F |

Table 2: Demographic variables (FFH = Fisher–Freeman–Halton exact test; F = Fisher's exact test)

#### 4.4.2 Negative comments (occasional to common)

Some criticized the predictability of the plot, stylistic repetition or emotional flatness. A few were clearly frustrated by tropes (e.g., the female temptress) or found the story preachy and uninspired.

#### 4.4.3 Neutral comments (rare)

A single note mentioned similarity with one of the other short stories in the experiment, without a clear judgment.

### 4.5 Author or text recognized

When reading the Oval story, one reader was reminded of an English author, "the one who wrote about a boy obsessed with Tony Hawk who gets his girlfriend pregnant." The description fits Nick Hornby. Cesare Pavese was named as the possible author of the Hexagon text, and both Cesare Pavese and Italo Calvino were associated with the Star text.

### 4.6 Demographic data

Despite clear instructions, only fifteen participants completed the demographic and reading-habits questionnaire. The respondents were divided into two roughly equal groups based on their ratings of the Oval story on a 0–10 scale (0 = strong dislike, 10 = strong appreciation). Scores below 6 were classified as low (n = 7), and scores of 6 or above as high (n = 8). These groups were then crossed with the categorical bands defined in the following subsections for each demographic variable, and contingency tables were built for each of the three stories. On the basis of these tables, as shown in Table 2, no statistically significant associations were found between any demographic variables and the scores assigned to the short stories ($p > .05$).

#### 4.6.1 Age group

The participants were grouped into three 20-year age bands.

#### 4.6.2 Gender

All the women assigned high preference scores to the star text, compared with only half of the men; this difference approached statistical significance (Fisher's exact $p = .057$). Neither of the other two texts came close to showing a gender association.

#### 4.6.3 Educational level

The various qualifications reported were grouped into lower secondary, upper secondary and tertiary level.

#### 4.6.4 First language

All the participants reported Italian as their first language, with three exceptions: two bilingual speakers (Italian–Romanian and Italian–French) and one native Spanish speaker. For the purposes of analysis, first language was categorized as Italian versus non-Italian/bilingual. Although both bilingual participants and the native Spanish speaker assigned high scores to the Oval and Star texts, the small size of this group prevented this pattern from reaching statistical significance.

### 4.7 Reading habits

The same preference rating categories defined for the demographic data were applied to the reading-habits data. As shown in Table 3, no statistically significant associations were found between reading-habit variables and the scores assigned to the short stories ($p > .05$).

| Reading habit | Oval (p value / test) | Hexagon (p value / test) | Star (p value / test) |
|---|---|---|---|
| **Reading frequency** | 1.000 / F | 1.000 / F | .476 / F |
| **Typical reading** | 1.000 / FFH | 1.000 / FFH | 1.000 / FFH |
| **Preferred reading** | .782 / FFH | 1.000 / FFH | .352 / FFH |

Table 3: Reading-habit variables (FFH = Fisher–Freeman–Halton exact test; F = Fisher's exact test)

### 4.7.1 Reading frequency

The participants rated their reading frequency on a 1–10 scale (1 = rarely, 10 = constantly). Since the lowest score reported was 5, responses were grouped into moderate (5–7) and high (8–10) frequency for the purposes of analysis.

### 4.7.2 Typical reading

The participants were asked about the kinds of materials they typically read. Given the limited sample size, the results were grouped into three mutually exclusive categories: fiction, non-fiction and both.

### 4.7.3 Preferred reading

The participants were also asked what they would prefer to read if given free choice. When no preference was stated, typical reading was used as a proxy, and responses were categorized using the same three groups.

## 5 Discussion

The experiment took place in the researcher's hometown (just under 15,750 inhabitants). Consequently, some of the participants knew him, and in at least two cases – but probably only two – were aware of his research into AI-generated texts. One of them noted on her feedback form that she believed some of the stories were AI-generated and correctly identified both SynTs. However, since she still gave them medium to high scores (5 and 7), her data was retained. No other participants mentioned AI in their feedback.

The mean scores assigned to each text fall within a narrow range: the HAT (Oval) received an average score of 6.83, Hexagon 7.33 and Star 7.42, all out of 10. While the SynTs slightly outperformed the human-authored one on average, the differences were modest. In terms of reader preference, Hexagon was most frequently ranked first (9 times), followed by Star (7 times) and Oval (6 times). These outcomes suggest no clear disadvantage for AI-generated Italian fiction under blind conditions.

A scatterplot comparing human mean scores with Plagramme scores for the three texts (Chart 1) shows that while the human scores are tightly clustered, the AI-detector scores vary widely, from 17% to 94%. Consequently, there is no real alignment or consistent relationship between the two scores.

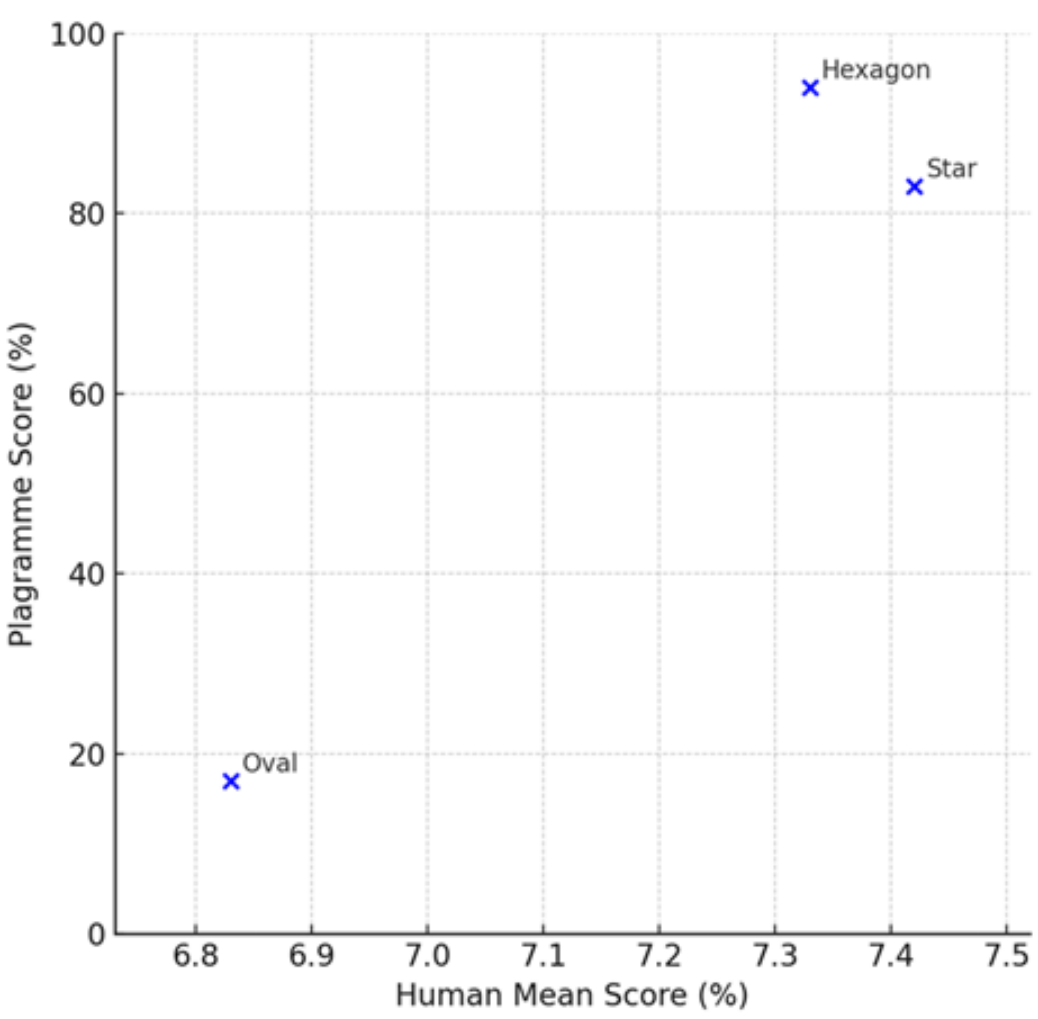

Chart 1: Scatterplot comparing human mean scores with Plagramme scores

In the qualitative comments, the HAT (Oval) was frequently praised for its narrative voice, humour and character development, though some criticized its pacing and verbosity. Hexagon was described as fluid and emotionally engaging but occasionally clichéd or overly moralistic. Star was noted for its immersive style and psychological insight yet drew some criticism for predictability and stylistic repetition. Despite these varied evaluations, no text received unequivocal acclaim or rejection. The balance of praise and criticism suggests that all three texts were assessed primarily on their perceived literary and emotional qualities.

The SynTs were associated with Cesare Pavese, and in one case also with Italo Calvino, who were contemporaries and editorial colleagues at Einaudi. Calvino's early neorealist work shares affinities with both Pavese and Moravia, particularly in its portrayal of wartime and youth. Interestingly, Pavese was linked to both AI-generated texts, perhaps due to a perceived tone of introspection, existential reflection or emotional restraint. The fact that the real Italian story was instead associated with a contemporary English novelist, Nick Hornby, suggests that the AI imitations may have appeared more "literary" or stereotypically Italian than the authentic text.

None of the readers in this study commented on the grammatical errors or the English syntactic and lexical influence[6] observed in Farrell's previous experiments with the same texts (2025a; 2026). This difference may be attributed to the task: here, participants were simply asked to indicate their preferred text, whereas in the earlier studies they were explicitly asked to identify AI-generated writing. It may also suggest that the professional and trainee translators, who took part in the earlier experiments, pay attention to different textual features than casual readers, whose focus may be more on content and emotional resonance than on linguistic accuracy.

Regarding first language, it is interesting to observe that two participants identified as bilingual. In Farrell's earlier study involving 69 professional translators (2026), none identified as bilingual. This may reflect a difference in how bilingualism is interpreted. While professional translators clearly speak more than one language, when asked to report their first language, they typically indicate their dominant one.

As reported in the results, both bilingual participants and the native Spanish speaker fell within the high-score group when assessing the Oval text (HAT), although this difference was not statistically significant. This may suggest a subtle trend. One possible explanation might lie in the calques, semantic loans and syntactic transfer from English observed in the SynTs (Farrell, 2026). It may be that bilingual people and native Spanish speakers are more appreciative of the distinctive features of an authentic Italian text. At the same time, however, this interpretation is contradicted by the high scores they also gave to the Star text.

There appears to be no clear explanation for the possible association between gender and preference for the Star story, which, in any case, did not reach the conventional threshold for significance ($p < .05$). No other demographic or reading-habit variables approached significance, suggesting they are unlikely to have influenced preferences among the texts used in this experiment.

### 5.1 Comparison with the results of the previous experiments

As can be seen in Table 4, the HAT received the lowest preference rating from the readers and was rated as most likely to be AI-generated by the postgraduates in Farrell's previous study (2025a). In contrast, one of the AI-generated texts (Star) received the highest preference rating and was judged by both the postgraduates and professional translators to be the most *human-like* (Farrell, 2025a; 2026). The postgraduates' AI-likeness scores exhibit a perfect inverse correlation with the reader's preference rankings (Spearman's $\rho = -1.0$), while the correlation between reader and translator scores is moderately negative ($\rho = -0.5$, not statistically significant), and the correlation between postgraduates and translators is moderately positive ($\rho = 0.5$). These patterns suggest a potential alignment between reader preference and the perception of human authorship, regardless of whether the text was in fact human-written. However, the small sample size (only three texts) limits the statistical power, and the correlations – while mathematically strong – may reflect chance alignments rather than robust trends.

### 5.2 Need for synthetic-text editing

The rationale behind STE rests on two key assumptions: first, that readers can reliably distinguish between SynT and HAT; and second, that they prefer reading HAT.

Regarding the first assumption, one participant – whose identity was revealed indirectly through the content of her feedback – explicitly identified the AI-generated texts, suggesting that detection is possible. This participant taught French at a university in Milan and regularly assessed student work for unauthorized AI use. Her background may have contributed to her ability to detect the stylistic and structural cues typical of synthetic writing. This observation, along with findings from previous studies (Dou et al., 2022; Farrell, 2025a; 2026), indicates that some people are probably able to identify AI-generated texts, particularly in professional or evaluative contexts.

| Text | Kind | Reader ranking | Postgrad ranking | Translator ranking | AI detector ranking |
|---|---|---|---|---|---|
| **Oval** | HAT | 3 | 1 | 2 | 3 |
| **Hexagon** | SynT | 2 | 2 | 1 | 1 |
| **Star** | SynT | 1 | 3 | 3 | 2 |

Table 4: Comparison of rankings from all three experiments and the AI detector

[6] Although it may seem surprising to mention English syntactic and lexical influence in the context of AI-generated Italian text rather than machine translation, Farrell (2026) hypothesizes that the observed effect may result from the well-documented predominance of English-language data in ChatGPT's training.

The findings of this study challenge the second assumption. Both AI-generated texts received higher mean scores and more first-place rankings than the HAT, and the distribution of reader praise and criticism was relatively balanced across all three. These results are in line with those of Farrell's second study, which also found that some readers may actually prefer SynT when authorship is not disclosed.

The textual anomalies observed in the Italian SynTs during Farrell's earlier experiments (2025a, 2026) included low burstiness (the natural variation in sentence length and rhythm typical of human writing), as well as narrative contradictions and continuity errors. Similar issues have been reported in English SynTs (Dou et al., 2022). Farrell also noted influences from English syntax and lexical calques, along with Italian grammatical errors. However, this study shows that casual readers do not appear to readily detect these faults. Moreover, grammar tends to be less of an issue in English SynT, which exhibits a high degree of grammatical and orthographic accuracy (Radtke and Rummel, 2025).

In light of these findings, STE may not be critical for the kinds of texts analysed in this study. This does not imply, however, that STE is unnecessary in high-stakes or accuracy-sensitive domains, such as medicine, law or science, where textual errors could have serious consequences. It may also function much like the preference for handmade over mass-produced products: driven not by necessity, but by a desire for authenticity.

### 5.3 Limitations

This experiment was limited to a small selection of texts of a similar kind in a single language and was carried out on a small number of people. As a result, the findings and conclusions of this study may not be broadly generalizable.

## 6 Conclusion

The findings of this experiment suggest that, in blind conditions, readers may prefer AI-generated short stories in Italian to those written by established human authors. The slightly higher ratings and first-place rankings for the SynTs indicate that such narratives can meet or even exceed the expectations of casual readers in terms of engagement, clarity and emotional resonance. Notably, the specific SynTs chosen for this study were the ones most often mistaken for human-written in previous experiments. In other words, we seem to like what we perceive to be human, even when that perception is mistaken. No demographic or reading-habit factors were found to influence these preferences significantly.

While STE may remain important in high-stakes or technical domains, its necessity for creative fiction – at least in Italian – appears less clear-cut. Nevertheless, given the small number of texts, these interpretations should be treated as tentative. Further study with more material and a broader reader base is needed to validate these observations and explore the robustness of the relationship between preference and perceived – though not necessarily actual – human authorship.

## Acknowledgments

The author gratefully acknowledges the support of the staff at the *Biblioteca Civica 'F. Pezza', Civico.17*, in Mortara, Pavia, Italy, as well as the volunteer participants whose contributions made this study possible.